\documentclass{article}

\usepackage{microtype}
\usepackage{graphicx}
\usepackage{subcaption}
\usepackage{booktabs} 
\usepackage{makecell}
\usepackage{tablefootnote}

\usepackage{hyperref}

\usepackage[accepted]{icml2026}

\usepackage{amsmath}
\usepackage{amssymb}
\usepackage{mathtools}
\usepackage{amsthm}
\usepackage{multirow}
\usepackage{multicol}
\usepackage[most]{tcolorbox}
\usepackage[version=4]{mhchem}

\newcommand{\HNMR}{$^{1}$H-NMR }
\newcommand{\CNMR}{$^{13}$C-NMR }

\newcommand{\toolcall}[1]{\par\smallskip\noindent{\ttfamily\small$\hookrightarrow$\ tool call: #1}\par\smallskip}

\usepackage[capitalize,noabbrev]{cleveref}

\theoremstyle{plain}

\theoremstyle{definition}

\theoremstyle{remark}

\usepackage[textsize=tiny]{todonotes}

\icmltitlerunning{\smash{NMR Elucidation as Agentic Search}}

\begin{document}

\twocolumn[
  \icmltitle{NMR Elucidation as an Agentic Search Problem, Not a Modeling Problem}


  \icmlsetsymbol{equal}{*}

  \begin{icmlauthorlist}
    \icmlauthor{Irina Espejo Morales}{equal,1,2}
    \icmlauthor{Damon Hinz}{equal,1,2}
    \icmlauthor{Marvin Alberts}{3}
    \icmlauthor{Geraud Krawezik}{1,4}
    \icmlauthor{Haewon Jeong}{1,5,6}
    \icmlauthor{Shirley Ho}{1,2,7,8,9}

  \end{icmlauthorlist}

  \icmlaffiliation{1}{PolymathicAI}
  \icmlaffiliation{2}{Center for Data Science, New York University}
  \icmlaffiliation{3}{Independent}
  \icmlaffiliation{4}{Scientific Computing Core, Flatiron Institute}
  \icmlaffiliation{5}{Center for Computational Mathematics, Flatiron Institute}
  \icmlaffiliation{6}{University of California, Santa Barbara}
  \icmlaffiliation{7}{Center for Computational Astrophysics, Flatiron Institute}
  \icmlaffiliation{8}{Department of Astrophysical Sciences, Princeton University}
  \icmlaffiliation{9}{Department of Physics, New York University}

  \icmlcorrespondingauthor{Irina Espejo}{iem244@nyu.edu}
  \icmlcorrespondingauthor{Shirley Ho}{shirleyho@flatironinstitute.org}

  \icmlkeywords{Machine Learning, ICML}

  \vskip 0.3in
]



\printAffiliationsAndNotice{
\icmlEqualContribution}

\begin{abstract}
Structural elucidation from Nuclear Magnetic Resonance (NMR) data remains a fundamental bottleneck across chemistry, materials science, and biology. We demonstrate that an agentic AI system can perform this task at a level comparable to graduate-level chemistry students. Instead of training a model to directly map spectra to structures, we build a single autonomous agent, backed by a frozen LLM, that interacts with a curated environment with access to domain-specific processing tools, validation checks, tabulated chemical shifts, and instructions that outline the stepwise nature of a chemist’s thinking process. On the Alberts dataset, our agent elucidates structures with a top-1 accuracy of 71\%, comparable to the performance of graduate students at 66\% top-1 accuracy. On the van Bramer and AstraZeneca datasets, our agent achieved 80\% and 20\% top-1 accuracy respectively, outperforming zero-shot end-to-end deep learning models which were trained on large datasets of simulated spectra. These results show that reframing NMR elucidation as an LLM-guided constrained search, rather than a modeling task, yields substantial gains and suggests a path toward multi-step orchestration frameworks that integrate a variety of tools, models, and domain knowledge to assist in automating spectroscopic analysis. 

\end{abstract}


    \begin{figure*}[ht]
        \centering
        \includegraphics[width=\linewidth]{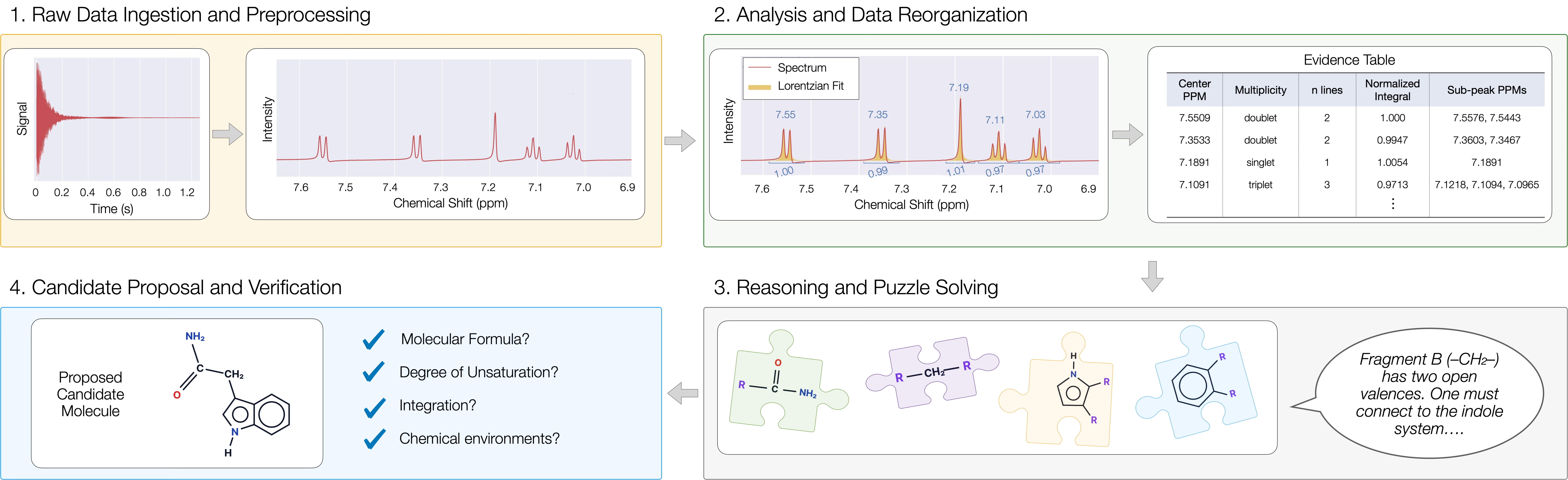}
        \caption{\textbf{Overview of NMR data pipeline for proposing molecular candidates}. The proposal pipeline is broken into four different phases: (1) Data Ingestion and processing - Ingesting the raw data and transforming it into human readable forms. (2) Analysis and data organization - Perform peak deconvolution and analysis of peaks, then generate data tables to use for reasoning about the data. (3) Reasoning - Reason on the extracted data, find correlations between peaks and what substructures may produce their signals. (4) Proposal and verification - Propose a molecule and verify that the molecule fits within the constraints of the data.}
    \end{figure*}
    
\section{Introduction}

Determining the structure of an unknown organic compound from routine spectra is a central task in analytical chemistry, underpinning drug discovery, natural product characterization, metabolomics, and quality control. However, the process is time consuming and requires years of experience. A chemist approaches the problem like a puzzle: given spectra and prior knowledge such as a molecular formula or a synthetic route, they derive constraints and reason over candidate molecules that satisfy them~\cite{breitmayer_eberhard_structure_2002}.

The earliest attempt to automate structure elucidation was DENDRAL~\citep{dendral,lederberg_how_1987}, one of the first expert systems designed to generate candidate chemical structures from raw Mass Spectrometry (MS) data. DENDRAL encoded a large knowledge base about MS, chemistry, and graph theory, which it leveraged to search for plausible structures and learn pruning rules that reduced the candidate space. The system's limitation was inherent to its design as all knowledge had to be hand-coded by domain experts as explicit rules, making the system brittle and difficult to generalize. The field subsequently shifted away from constraint-based search~\cite{steinbeck_seneca_2001,funatsu_recent_1996,munk_computer-based_1998,howarth2020dp4ai}. 

With the rise of machine learning, the dominant paradigm has become end-to-end prediction. A model is trained to directly map a spectrum to a molecular structure, typically represented as a SMILES string. This approach has seen substantial progress across all major spectroscopic modalities with models starting to achieve human level accuracy and revealing insights previously not possible~\cite{jonas_deep_2019,alberts2023LearningNMR,hu2024multitasknmr,stravs_msnovelist_2022,bohde_diffms_2025,wu_transformer-based_2025,alberts2025IRBenchmark,priessner2026mmst,alberts_automated_2025,wang2025nmrmind, jin2025nmrsolver}. However, these direct approaches lack the interpretability and explicit reasoning of earlier expert systems.

The original puzzle-solving paradigm has received comparatively little attention in the modern machine learning era. Only recently have multimodal LLMs been evaluated on structure elucidation, framed explicitly as a puzzle-solving task requiring iterative hypothesis testing and integration of multifaceted spectroscopic data~\cite{guocan2024molpuzzle}. Concurrent work on IR spectroscopy has explored multi-agent LLM frameworks that decompose the elucidation task into modular reasoning steps~\cite{yao2026iragent}. Yet no existing work has applied a fully agentic, constraint-based system to structure elucidation from one-dimensional NMR spectra, one of the most accessible and routinely collected spectroscopic data in organic chemistry.

In this paper, we revisit the puzzle solving approach: rather than learning a direct spectrum-to-structure mapping, we explore whether an agent can apply the puzzle-solving strategy on real world spectroscopic data. Our approach uses raw experimental NMR spectra together with the molecular formula, constructs structural constraints from the spectra, generates candidate molecules autonomously, and verifies them against the observed data without human intervention.

Our contributions are listed as follows:
\begin{itemize}
\item \textbf{Real experimental inputs.} We operate directly on raw NMR instrumentation files rather than tabulated peaks, reflecting the most realistic lab deployment setting.

\item \textbf{Agentic framework and environment.} We evaluate modern agentic systems for NMR elucidation using only 1D NMR data and molecular formula as input. We introduce an agent-based harness for context management, structured reasoning, and preprocessing, together with preprocessing tools independent of commercial software.

\item \textbf{Training-free approach.} Our approach does not involve training or the simulation of spectra datasets; instead it uses a frozen LLM as the backbone model and shifts the focus from assembling large datasets and leveraging compute to building an environment that distills the workflow of a human chemist.

\item \textbf{Autonomous search.} The agent generates candidate molecules from scratch without prior seed structures and handles diverse solvent conditions.

\item \textbf{Self-verification.} We designed validation tools that allow the agent to check candidate consistency against observed data without access to ground truth.
\end{itemize}

\section{Methods}
Our workflow starts with \texttt{FID} files (Free Induction Decay) which are preprocessed in a deterministic manner producing partially processed peak lists. For each task, the agent receives the same type of formats for inputs and outputs. \textbf{Inputs}: Partially processed peak lists. \textbf{Outputs}: SMILES strings of the top-10 candidates. The agent is able to adjust the partially processed peak lists as needed (remove noise, re-normalize, etc.) similar to how a human experimenter would during the reasoning process.

\subsection{Data Preprocessing}
    Before the agent accesses the raw files, all raw FID and acquisition data are sanitized to remove any identifying metadata that could reveal the answer to the agent. The system can ingest raw NMR data collected using Bruker, Varian, and JEOL instruments. The agent is provided with an automated data processing pipeline whose exact output format of the peak lists depends on the type of NMR modality. For \HNMR, the pipeline returns a list of the non-normalized peak positions, the normalized peak positions, the solvent peak position that was used for normalization, the integration ratios, and an initial attempt to normalize the peak integrations (the agent has autonomy to adjust the integrations or re-reference the spectrum to a different peak). For \CNMR, the processing pipeline returns the non-normalized peak positions, the normalized peak positions, and the solvent peak position used for normalization. The steps of the NMR processing are further explained below in the following sections.

    \subsubsection{FID Preprocessing}
        The FID first must be ingested and preprocessed prior to conversion to the frequency domain. For all vendor data, this includes apodization using exponential multiplication line broadening to improve the signal-to-noise ratio, and zero-filling to increase frequency resolution. In the case of Bruker data, the digital filter is removed and any trailing bytes are also truncated to the nearest 4-byte boundary. Other than the trailing bytes truncation, all of these processing steps are performed using functionality provided by the \texttt{nmrglue} Python library \citep{helmus_nmrglue_2013}.

    \subsubsection{Spectral Processing}
        A Fast Fourier Transform (FFT) is performed on the preprocessed FID to convert it from the time domain into the frequency domain, which is then converted to the ppm scale. This is then followed by phase and baseline correction. The phase correction first attempts to use the ACME autophase algorithm, as implemented in \texttt{nmrglue} \citep{chen_acme_autophase_2002}. This phasing is then scored, and if it falls under a threshold score, a global grid search for the best phasing parameters is performed to maximize the phase score. 
        
        \begin{equation}
            \label{eq:phase_score}
            \text{phase score} = \frac{\sum_{i \in S,\ x_i > 0}x_i}{\sum_{i \in S, \ x_i < 0} |x_i|}
        \end{equation}

        Where $x_i = \text{Re}(\phi(spectrum,\ p_0,\ p_1))$ is the real part of the spectrum phase ($\phi$) being maximized after applying zero-order and first-order phase corrections, and $S$ is the length of a signal mask, restricting the scoring to regions that actually contain peaks so noise doesn't distort the metric.
        The baseline correction is performed using an adaptive median-filter baseline correction. The solvent/TMS peak is then determined and the ppm scale is normalized to give the correct chemical shifts. Peak picking is then performed using the mean absolute deviation of the spectrum as a noise filter for SciPy's peak picking algorithm \citep{SciPy}. Finally, for \textsuperscript{1}H NMR, Lorentzian peaks are fitted to each peak following:

        \begin{equation}
            \label{eq:lorentzian}
            L(x) = \frac{A\gamma^2}{(x-x_0)^2 + \gamma^2}
        \end{equation} 

        \noindent where $L(x)$ is the Lorentzian peak, $x_0$ is the position, $A$ is the amplitude, and $\gamma$ is the half width at half maximum (HWHM). These peaks can then be used by the agent to begin building evidence tables for elucidation of the molecular species.

\subsection{Agentic System}

The agent is a single LLM driver built upon OpenAI AgentsSDK with access to a registry of curated tools but without coding capabilities and without browsing access. The prompt for the agent is assembled at runtime using a set of knowledge documents which provide all of the data input/output formats, tool documentation, data tables, etc. These knowledge documents provide no examples of NMR elucidation problems, only general chemistry knowledge about NMR elucidation and a description of how to approach structural elucidation. Crucially, the agent is given no seed structures and no access to the wider internet to ensure that any generated structures are due to its own reasoning, or information stored within the model. The agent only has access to the raw NMR spectra and the molecular formula from which it is tasked to predict the correct structure.

Once the data preprocessing is complete, the agent is given a dictionary called the \texttt{EvidenceTable}, which contains lists of peaks which are very roughly filtered, to remove clear baseline noise, and imprecisely processed (ppm, integration, etc.) The agent is allowed to reprocess the peaks if it feels it needs to, and adjust the table accordingly. From this table, the agent first generates an \texttt{EnvironmentTable} where it tries to classify different functional groups and substructures from the spectra. The agent then uses the \texttt{EnvironmentTable} to assemble different possible structures for proposal, using its internal knowledge to reason about the peak splitting, J-couplings, etc. searching for correlated peaks to justify the structure proposals.

\subsection{Verification Rules}
    In order for the agent to be able to determine a stopping point for the proposed structure, it is given a set of tools to verify that the molecule fits the data. It must be emphasized that this is \textit{not} verifying that the prediction matches the ground truth, but rather that the prediction fits the data constraints, ensuring that the model is predicting a structure that aligns with the data it is provided.

    \textbf{Molecular Formula:} It checks that the formula fo the candidate proposed by the agent matches the input formula exactly. This is a hard constraint that the agent must satisfy. 

    \textbf{Degree of Unsaturation (DoU):} This quantity is calculated symbolically and counts the number of rings and double/triple bonds present in a molecule. The rule checks that the integer value of the DoU stored in the evidence table against the value computed from the molecular formula input. It ensures that when constructing the evidence table no drift in the DoU value occurs. 

    \textbf{Integration Values (\textsuperscript{1}H-NMR):} It checks that the observed \textsuperscript{1}H integrals account for (approximately) every proton the formula asks for. The tolerance is $\pm$1 protons to account for noise, as well as a tolerance given for the \textit{possible} number of exchangeable protons given by the sum of number of nitrogen, oxygen, and sulfur atoms in the molecular formula.

    \textbf{Chemical Environment Counts:} The last constraint we check the chemical environments which compares the number of distinct \textsuperscript{1}H and \textsuperscript{13}C environments in the evidence table against the number of expected environments based on the topological symmetry of the proposed molecule which is calculated using \texttt{RDKit}.

\section{Experiments}

\subsection{Evaluation Datasets}
All datasets used for evaluation are experimental, and we use the raw instrument files as input without any human preprocessing or intervention. We deliberately exclude peak-picked tabulated reports extracted from chemistry papers. Because these reports are human generated, they vary by practitioner and introduce biases that do not reflect real-world inference conditions.

\paragraph{Chemistry Education~\citep{chem_edu}.} Consists of 247 small organic molecular structures paired with their \HNMR and \CNMR NMR spectra \texttt{mnova} files that have been curated for educational purposes. It contains generally clean spectra, with clean peak separation for low complexity structures.

\paragraph{Alberts Dataset~\citep{alberts_automated_2025}.} As proposed, this dataset originally consists of 16 small organic molecules. One of the samples was found to be degraded and we removed this sample from our evaluation. As such, we evaluate on 15 molecules instead of 16, and re-adjusted the accuracies reported in ~\citep{alberts_automated_2025} to reflect this reduction in the dataset. For each molecule \HNMR, \CNMR and IR spectra were measured. Human experts analyzed the spectra for each molecule, providing a strong human-performance baseline. As one of the molecules in the dataset is degraded and does not match the measured spectra, we exclude this molecule from our evaluation and scale the benchmark results accordingly. 

\paragraph{AstraZeneca Dataset~\citep{priessner2026mmst, astrazeneca2}.} This dataset consists of 34 molecules for which the \HNMR and \CNMR were measured, among others. The molecules are chemically diverse, drug-like structures, selected to assess model performance under realistic laboratory conditions, also accounting for real-world factors such as impurities and spectral noise.

\begin{table*}[!ht]
\caption{\textbf{Benchmark results on experimental datasets for NMR elucidation.} We evaluate molecular structure prediction from experimental spectra and molecular formula on three datasets: Chemistry Education (van Bramer), Alberts, and AstraZeneca. We report top-$k$ accuracy and Tanimoto similarity to the ground-truth structure (higher is better). Column $N$ indicates the total number of samples used by a given method for metrics calculations. Results for prior work and human evaluations are extracted from the original sources. See notes below for differences in evaluation protocols.} 
\label{tab:combined_benchmarks}
\centering
\renewcommand{\arraystretch}{1.2}
\setlength{\tabcolsep}{3pt}
\footnotesize

\begin{tabular}{llccccccc}
\toprule
 &  & \multicolumn{2}{c}{Top--1} & \multicolumn{2}{c}{Top--2} & \multicolumn{2}{c}{Top--5} &  \\
\cmidrule(lr){3-4}\cmidrule(lr){5-6}\cmidrule(lr){7-8}
 & \textsc{Method} & Acc.\% $\uparrow$ & Tan. $\uparrow$ & Acc.\% $\uparrow$ & Tan. $\uparrow$ & Acc.\% $\uparrow$ & Tan. $\uparrow$ & $N$ \\

\midrule
\multicolumn{9}{c}{\textbf{van Bramer \cite{chem_edu}}}\\
\midrule

\multirow{4}{*}{Model} & Alberts et. al. (zero shot)$^\dagger$ & 50.30 & N/A & 64.00 & N/A & 67.10 & N/A & \multirow{3}{*}{171} \\
 & Alberts et. al. (finetuned 5x CV)$^\dagger$ & 69.10{\scriptsize$\pm$11.20} & N/A & N/A & N/A & 91.50{\scriptsize$\pm$5.60} & N/A &  \\
 & \makecell[l]{Alberts et. al. (finetuned 5x CV + \\33k unpaired exp. samples)$^\dagger$} & \textbf{96.20{\scriptsize$\pm$7.50}} & N/A & N/A & N/A & \textbf{98.80{\scriptsize$\pm$2.50}} & N/A &  \\
\cmidrule(lr){2-9}
 & ChefNMR L (zero shot)$^{\dagger, *}$  & 56.00 & $\sim$0.68 & $\sim$65 & $\sim$0.75 & $\sim$70 & $\sim$0.80 & 238 \\
\cmidrule(lr){1-9}
\multirow{5}{*}{Agent} & Ours (\texttt{gpt-5.4}) & 80.42 & 0.88 & 84.58 & 0.90 & 88.33 & 0.93 & \multirow{3}{*}{236} \\
 & Ours (\texttt{kimi-k2.6}) & 80.87 & 0.90 & 84.35 & 0.90 & 90.00 & 0.93 &  \\
 & Ours (\texttt{qwen3.5-122b}) & 54.96 & 0.69 & 63.64 & 0.76 & 69.83 & 0.82 &  \\
\cmidrule(lr){2-9}
 & Baseline (\texttt{Codex 5.5}) & 62.30{\scriptsize$\pm$4.10} & 0.73{\scriptsize$\pm$0.03} & 65.70{\scriptsize$\pm$3.40} & 0.77{\scriptsize$\pm$0.02} & 72.60{\scriptsize$\pm$2.90} & 0.83{\scriptsize$\pm$0.02} & \multirow{2}{*}{236} \\
 & Baseline (\texttt{CC Opus-4.7}) & 55.60{\scriptsize$\pm$29.50} & 0.63{\scriptsize$\pm$0.34} & 58.90{\scriptsize$\pm$31.30} & 0.66{\scriptsize$\pm$0.35} & 63.70{\scriptsize$\pm$33.10} & 0.69{\scriptsize$\pm$0.37} &  \\

\addlinespace[2.0em]
\midrule
\multicolumn{9}{c}{\textbf{Alberts \cite{alberts_automated_2025}}}\\
\midrule
\multirow{1}{*}{} & Grad Students al.$^\dagger$ & 66.67{\scriptsize$\pm$3.06} & N/A & 82.21{\scriptsize$\pm$3.06} & N/A & 84.45{\scriptsize$\pm$3.95} & N/A & 15 \\

\midrule
\multirow{1}{*}{Model} & Alberts et. al.$^\dagger$ & 69.33{\scriptsize$\pm$3.06} & N/A & 77.34{\scriptsize$\pm$3.06} & N/A & 81.33{\scriptsize$\pm$3.95} & N/A & 15 \\
\cmidrule(lr){1-9}
\multirow{5}{*}{Agent} & Ours (\texttt{gpt-5.4}) & 66.67{\scriptsize$\pm$5.44} & 0.76{\scriptsize$\pm$0.04} & 71.11{\scriptsize$\pm$3.14} & 0.80{\scriptsize$\pm$0.03} & 71.11{\scriptsize$\pm$3.14} & 0.80{\scriptsize$\pm$0.02} & \multirow{3}{*}{15} \\
 & Ours (\texttt{kimi-k2.6}) & \textbf{71.11{\scriptsize$\pm$6.29}} & \textbf{0.80{\scriptsize$\pm$0.04}} & \textbf{75.56{\scriptsize$\pm$3.14}} & \textbf{0.81{\scriptsize$\pm$0.03}} & \textbf{77.78{\scriptsize$\pm$3.14}} & \textbf{0.82{\scriptsize$\pm$0.03}} &  \\
 & Ours (\texttt{qwen3.5-122b}) & 35.56{\scriptsize$\pm$11.33} & 0.54{\scriptsize$\pm$0.08} & 37.78{\scriptsize$\pm$12.57} & 0.57{\scriptsize$\pm$0.08} & 44.44{\scriptsize$\pm$13.70} & 0.62{\scriptsize$\pm$0.07} &  \\
\cmidrule(lr){2-9}
 & Baseline (\texttt{Codex 5.5}) & 40.00{\scriptsize$\pm$13.30} & 0.56{\scriptsize$\pm$0.11} & 46.70{\scriptsize$\pm$11.50} & 0.59{\scriptsize$\pm$0.11} & 48.90{\scriptsize$\pm$10.20} & 0.61{\scriptsize$\pm$0.09} & \multirow{2}{*}{15} \\
 & Baseline (\texttt{CC Opus-4.7}) & 24.40{\scriptsize$\pm$3.80} & 0.39{\scriptsize$\pm$0.04} & 28.90{\scriptsize$\pm$3.80} & 0.42{\scriptsize$\pm$0.05} & 33.30{\scriptsize$\pm$6.70} & 0.49{\scriptsize$\pm$0.06} &  \\

\addlinespace[2.0em]
\midrule
\multicolumn{9}{c}{\textbf{AstraZeneca \cite{priessner2026mmst}}}\\
\midrule

\multirow{4}{*}{Model} & Alberts et. al.$^\dagger$ & 14.70{\scriptsize$\pm$4.20} & N/A & 18.80{\scriptsize$\pm$2.40} & N/A & 22.90{\scriptsize$\pm$2.20} & N/A & 34 \\
\cmidrule(lr){2-9}
 & MMST (base model)$^\dagger$ & 0.00 & N/A & 0.00 & N/A & 3.00 & N/A & \multirow{3}{*}{34} \\
 & MMST (trained on analogues)$^\dagger$  & 12.00 & N/A & 38.00 & N/A & 44.00 & N/A &  \\
 & MMST (trained on test)$^\dagger$  & 31.00 & N/A & 56.00 & N/A & 81.00 & N/A &  \\
\cmidrule(lr){1-9}
\multirow{5}{*}{Agent} & Ours (\texttt{gpt-5.4}) & 7.84{\scriptsize$\pm$1.39} & 0.38{\scriptsize$\pm$0.01} & 9.80{\scriptsize$\pm$1.39} & 0.39{\scriptsize$\pm$0.00} & 15.69{\scriptsize$\pm$1.39} & 0.41{\scriptsize$\pm$0.00} & \multirow{3}{*}{34} \\
 & Ours (\texttt{kimi-k2.6}) & \textbf{20.60{\scriptsize$\pm$0.49}} & \textbf{0.63{\scriptsize$\pm$0.01}} & \textbf{27.94{\scriptsize$\pm$3.12}} & \textbf{0.64{\scriptsize$\pm$0.01}} & \textbf{29.06{\scriptsize$\pm$4.10}} & \textbf{0.65{\scriptsize$\pm$0.02}} &  \\
 & Ours (\texttt{qwen3.5-122b}) & 1.96{\scriptsize$\pm$1.39} & 0.22{\scriptsize$\pm$0.01} & 2.94{\scriptsize$\pm$2.40} & 0.23{\scriptsize$\pm$0.01} & 2.94{\scriptsize$\pm$2.40} & 0.24{\scriptsize$\pm$0.01} &  \\
\cmidrule(lr){2-9}
 & Baseline (\texttt{Codex 5.5}) & 12.70{\scriptsize$\pm$4.50} & 0.39{\scriptsize$\pm$0.09} & 13.70{\scriptsize$\pm$6.10} & 0.40{\scriptsize$\pm$0.09} & 15.70{\scriptsize$\pm$6.80} & 0.41{\scriptsize$\pm$0.09} & \multirow{2}{*}{34} \\
 & Baseline (\texttt{CC Opus-4.7}) & 6.90{\scriptsize$\pm$4.50} & 0.26{\scriptsize$\pm$0.07} & 6.90{\scriptsize$\pm$4.50} & 0.26{\scriptsize$\pm$0.07} & 7.80{\scriptsize$\pm$3.40} & 0.28{\scriptsize$\pm$0.07} &  \\

\bottomrule

\end{tabular}

\vspace{0.3em}
\begin{minipage}{0.95\textwidth}
\centering
\footnotesize
$^\dagger$ Reported performance from source. \hspace{0.5em}
$^{*}$ Reports accuracy removing stereochemical features.\\ 
\end{minipage}

\end{table*}

\subsection{Evaluation Settings}
\label{sec:eval}
Benchmarking NMR Elucidation methods under identical conditions is challenging. There are several aspects that limit direct comparison. First, methods use different combinations of spectroscopic modalities and different input formats for training and inference. Second, because of this, preprocessing of samples is not consistent across methods; not all methods can be evaluated on any dataset or even be evaluated on the full dataset, as some samples can be invalid for certain methods. Finally, reported accuracy depends on how SMILES stereochemistry handled. If removed completely, accuracy can be higher than it should be, because enantiomers and diastereomers may be counted as correct. However, depending on the modalities used, it might not be physically possible to differentiate between enantiomers, which over-penalizes a method on a task it cannot resolve.

\paragraph{Specialized Models.} We benchmark against different types of deep learning models, multi-modal, language models, such as \citet{alberts_automated_2025}'s model and the MultiModalSpectralTransformer (MMST)~\citep{priessner2026llm_elucidation}. 
\citet{alberts_automated_2025}'s model is trained on the molecular formula, \HNMR, \CNMR, and IR spectra to predict SMILES~\cite{weininger_smiles_1988} as output. This model is pretrained on simulated data before being finetuned on experimental data. On the other hand, MMST is trained exclusively on simulated data and follows the same input output format as \citet{alberts_automated_2025}'s model. We also include a diffusion-based model ChefNMR~\citep{nmrchef} trained on the molecular formula, \HNMR, and \CNMR which predict the 3D atomic positions of a molecule.

\paragraph{Agent Baselines.} We instantiate a coding agent, Claude Code (\texttt{claude-opus-4.7}) and Codex (\texttt{gpt-5.5}), in an isolated Docker container with access to a Python interpreter but without internet browsing or our domain-specific tools. It receives the same inputs as our system, the raw spectral data (molecular formula, \HNMR and \CNMR \texttt{fid} files), together with a comprehensive prompt that specified the task and the reasoning steps, similar to our agent's prompt. This setup allows for comparison between the LLM's out of the box capabilities and the environment that we constructed for the task of NMR elucidation.
We sandboxed both agents, baseline and ours, to avoid a leakage of the answer. Due to resource constraints, we were able to perform three independent runs for Codex for all the datasets, but only one for Claude Code, thus we report mean and standard deviation for Codex and the mean of a single full run for Claude Code.

\paragraph{Metrics}
We report top-$k$ accuracy based on exact matches between the canonical SMILES strings of the candidates proposed by the agent and the ground truth. We retain all stereochemical information in our SMILES, so an incorrect stereochemical prediction results \emph{always} in a zero hit.
We did this because, while NMR is unable to decipher enantiomers under the most common NMR conditions, they can be determined using chiral solvents as the NMR solvent. Additionally, since the stereochemistry of diastereomers can be determined (which can be taken advantage of to determine the stereochemistry of an enantiomer), we felt it was best to be consistent and always take stereochemistry into account when evaluating whether the model correctly determined the molecule or not, as there are cases in which the enantiomer can be determined.
This is in contrast with some of the models we benchmark against, such as \citet{nmrchef}, where stereochemistry is removed from the SMILES strings. The top-$k$ Tanimoto similarity is calculated using Morgan fingerprints (radius 2, 2048 bits) between the candidates and the ground truth. Per-dataset metrics are averaged across all tasks and reported as mean and standard deviation over independent runs.

\begin{figure*}[!t]
    \centering
    \includegraphics[width=0.9\linewidth]{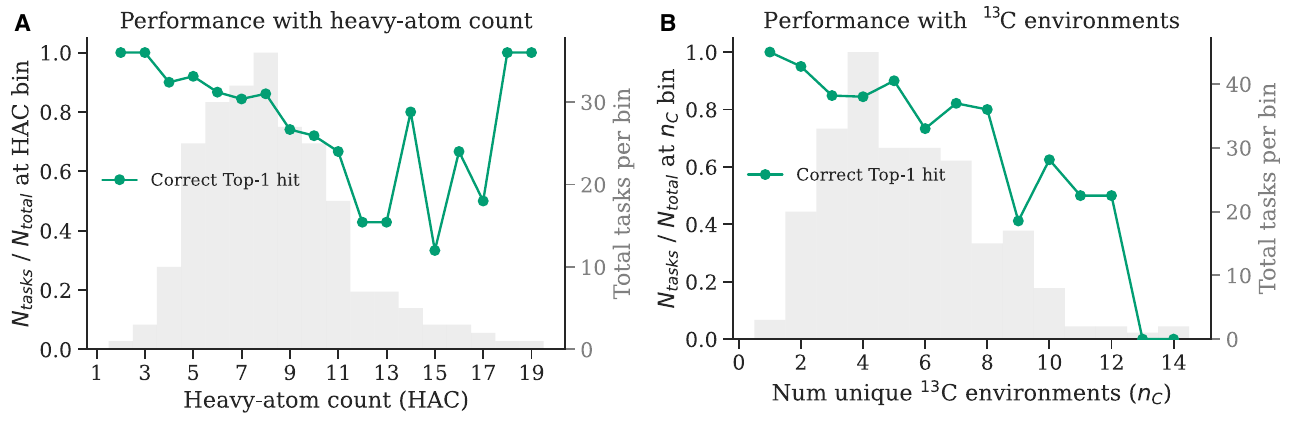}
\caption{\textbf{Stratified evaluations by selected molecular properties.} Performance on the Chemistry Education dataset for \texttt{kimi-k2.6} stratified by (A) Heavy-atom count and  (B) Number of unique $^{13}$C environments. Green lines show the fraction of tasks within a bin where the agent was correct. The faint gray histogram shows the total number of tasks per bin. Performance tends to decrease with the number of unique carbon environments, but so does the absolute number of tasks, making the extraction of conclusions more difficult.}
\label{fig:strat_evals}
\end{figure*}

\section{Results}
\subsection{Benchmarks}
The results on the Alberts, AstraZeneca, and Chemistry Education datasets are shown in \autoref{tab:combined_benchmarks}. These benchmarks cover complementary settings: expert-evaluation, drug-like molecules measured under realistic laboratory conditions, and cleaner educational spectra. We compare our agent with specialized NMR elucidation models, human experts where available, and general purpose coding-agent baselines. Because these methods differ in inputs, preprocessing requirements and metric reporting, the results should be interpreted together with the caveats described in~\Cref{sec:eval}. For the Chemistry Education evaluations, ChefNMR reports evaluations on 238/247 tasks due to inconsistencies or wrongly paired data, our Agent was able to complete 236/247 tasks, and \citet{alberts_automated_2025} evaluates on 171/247 in order to have paired 1D NMR and IR samples. For the Alberts and AstraZeneca datasets, all models are evaluated in the totality of the samples.

Across all datasets, our agent performs competitively with, and sometimes outperforms, models that have been trained on large amounts of simulated spectra. Regarding agentic baselines, the fact that a general purpose coding-agent consistently underperforms our agent indicates that the performance gains of our agent not only come from the underlying LLM, but from the task-specific environment that was provided. On the Alberts dataset, our agent has comparable performance to a graduate-level chemistry student. Irregardless of the method, performance degrades when testing on the more complex AstraZeneca dataset.

\subsection{Stratified evaluations}
In this section, we characterize stratified performance across different axes that make NMR elucidation more complex. For instance, it is common for models to degrade in performance with molecular weight, as shown in ~\citet{Alberts2024IRStructure} or when exposed to real-world molecules of interest, as shown in~\citet{alberts2024multimodal, nmrchef}. Additionally, as the number of unique chemical environments increases in the molecule, the complexity of the resulting spectrum increases and parts of the spectrum may become more difficult to interpret. One clear example of this is multiplet resolvability as peak overlap becomes more prevalent~\citep{aguilar_nmr_resolution_2010, halabalaki_natural_product_elucidation_2014}. \autoref{fig:strat_evals} shows the performance of the agent on the van Bramer dataset changes with heavy-atom count and the number of unique \textsuperscript{13}C environments, which can be regarded as a proxy for the molecular complexity and spectral complexity respectively. Unsurprisingly, as the complexity increases, the task of elucidating the structure becomes more difficult for the agent. 
For higher heavy atom counts and unique \textsuperscript{13}C environments the trends become relatively noisy. This can be attributed to the small number of molecules and spectra that fall in this range and is clearly an area which must be investigated in future studies.
\subsection{Case studies}

\begin{figure*}[t]
\centering
\begin{subfigure}{0.85\linewidth}
    \centering
    \includegraphics[width=\linewidth]{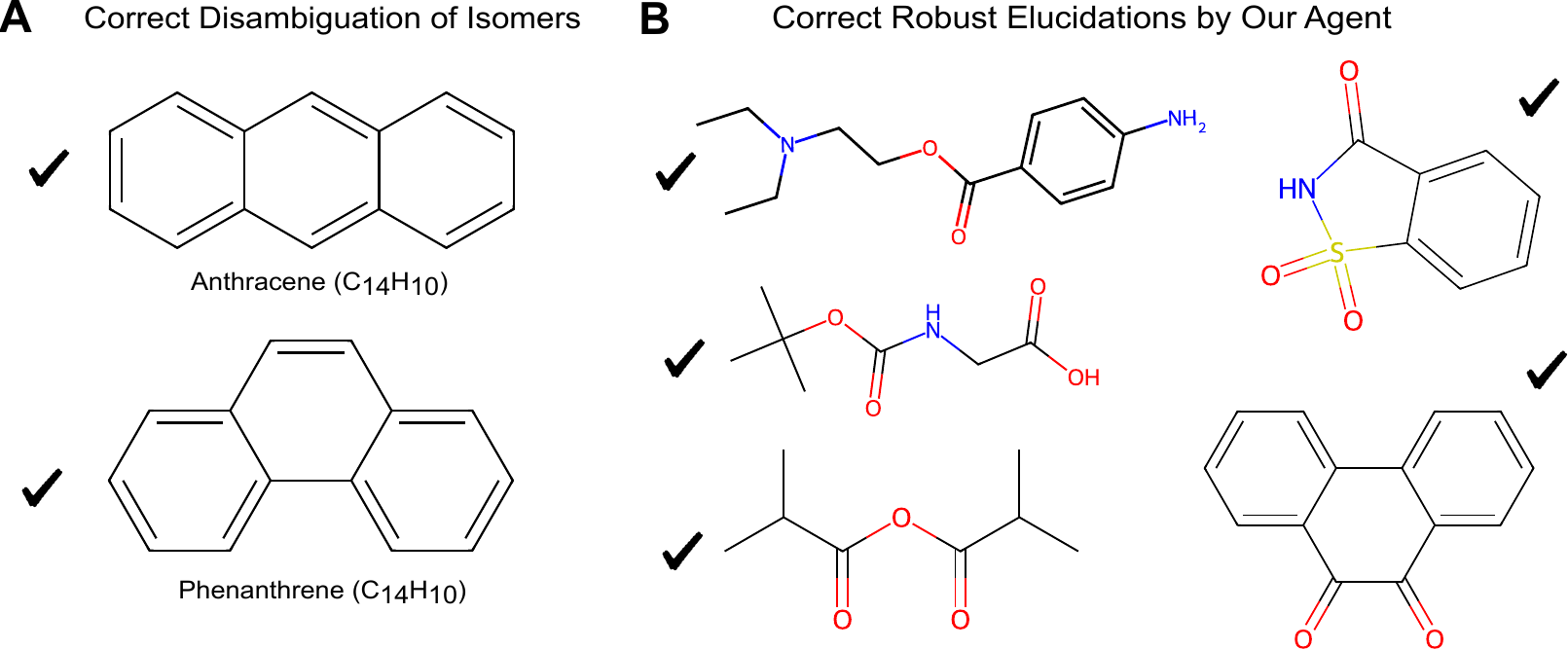}
    \label{fig:case_study_correct}
\end{subfigure}
\vspace{0.75em}
\begin{subfigure}{0.85\linewidth}
    \centering
    \includegraphics[width=\linewidth]{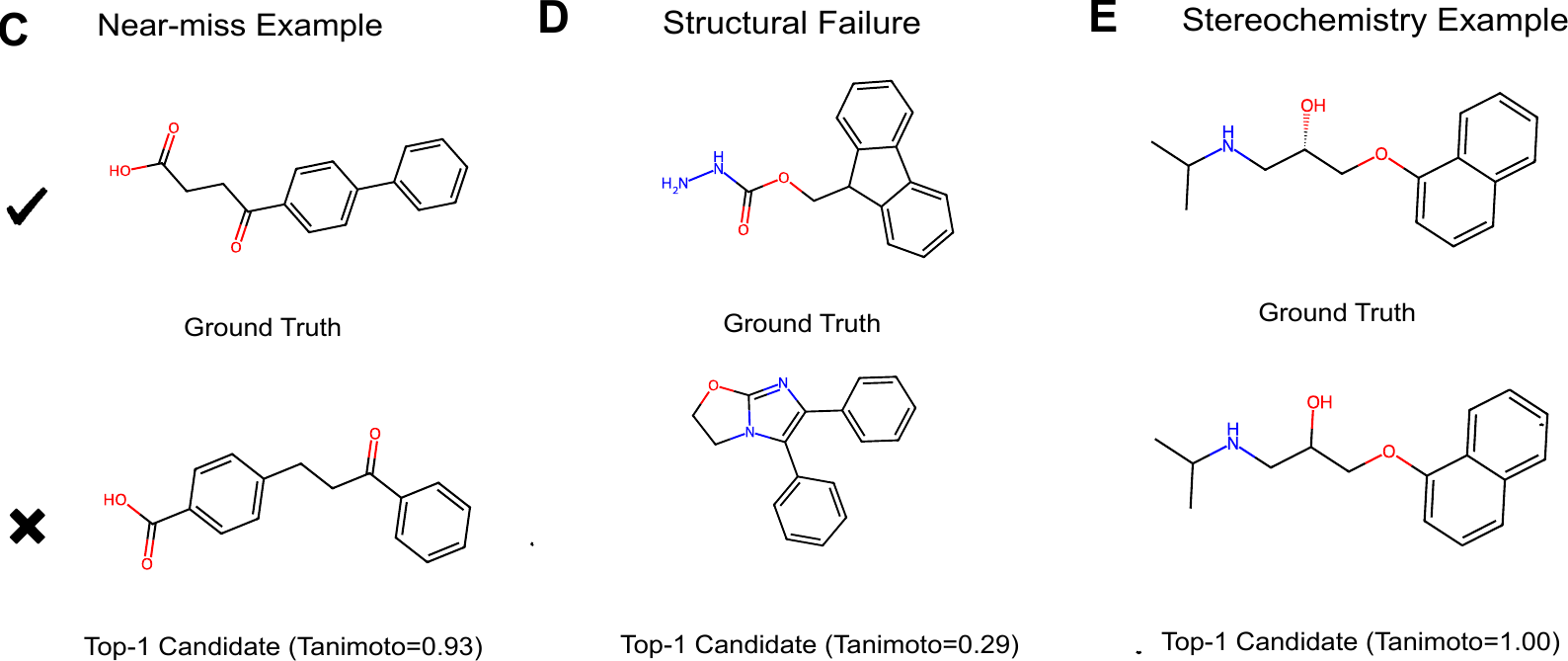}
    \label{fig:case_study_incorrect}
\end{subfigure}

\caption{\textbf{Case studies of top-1 predictions by our agent.} All examples are predictions from \texttt{kimi-k2.6}. Top: examples that were predicted robustly, with the agent returning the same answer in all or nearly all queries. (A) The agent correctly disambiguates two structurally similar isomers with the same molecular formula and never confuses them with each other. (B) Across a set of varied molecules, the agent predicts the correct structure in every trial. Bottom: representative failure modes. (C) A near-miss prediction with high structural similarity to the ground truth. (D) A structural failure with low similarity to the ground truth. (E) A stereochemical error, where the predicted connectivity is correct but the stereochemistry differs.}
\label{fig:case_study}
\end{figure*}

\autoref{fig:case_study} shows representative examples of successes and failures of our agent. 
We observe that the agent is able to differentiate between isomers, correctly interpreting the subtle differences between the spectra even in cases where isomerization would not lead to large changes in the spectra.
Three failure modes are prominent: near-misses, structural failures and stereochemistry-related failures. After analyzing the reasoning traces of the agent, we observed that structural failures are relatively varied. For instance, the model will often narrow in on a certain region of the chemical space, and struggles to re-consider earlier assumptions. When assigning spectral peaks as artifacts, noise, solvent or signal is more difficult, the agent can struggle to find a molecular candidate that satisfies the chemical formula, while also having accurate integrations and number of chemical environments, entering into a long reasoning loop that can derail. Some examples of successful and failed reasoning traces are given in the Supplementary Material.

\subsection{Run-time analysis}
\label{sec:runtime}
\autoref{tab:run_times} reports the per-run wall-clock time across model and benchmark groups. Run time scales with dataset difficulty: for both backbone models, the mean time grows from the cleaner Chemistry Education spectra to the expert-level Alberts molecules to the complex, drug-like AstraZeneca structures, mirroring the accuracy trends in \autoref{tab:combined_benchmarks} and consistent with the longer, harder-to-navigate reasoning traces discussed above. The choice of backbone dominates the cost: \texttt{kimi-k2.6} is roughly four to five times slower than \texttt{gpt-5.4} across all three datasets, and its larger standard deviations reflect occasional long reasoning loops on the harder tasks. In every group, the LLM accounts for the overwhelming majority of wall-clock time (83--99\%), so end-to-end latency is governed almost entirely by the frozen backbone rather than by the tool execution and preprocessing overhead of our environment.

\begin{table}[!htbp]
\caption{Per-run wall-clock timing across model/benchmark groups, 
over a total of 761 runs. Times shown as mean $\pm$ sample
standard deviation. The \emph{model\%} column is the fraction of total
wall-clock time spent in LLM (\texttt{response}/\texttt{generation}) spans versus
tool execution.} \label{tab:run_times}
\centering
\footnotesize
\setlength{\tabcolsep}{5pt}
\renewcommand{\arraystretch}{1.1}
\begin{tabular}{lrrr}
\toprule
Group & $n$ & Wall (\texttt{M:SS}) & model\% \\
\midrule
gpt-5.4/chem\_edu       & 238 & $4{:}38 \pm 3{:}11$   & 83\% \\
gpt-5.4/alberts         &  45 & $7{:}03 \pm 4{:}31$   & 96\% \\
gpt-5.4/astrazeneca     & 101 & $13{:}49 \pm 5{:}21$  & 92\% \\
\midrule
kimi-k2.6/chem\_edu     & 235 & $20{:}44 \pm 19{:}11$ & 96\% \\
kimi-k2.6/alberts       &  45 & $30{:}31 \pm 23{:}36$ & 99\% \\
kimi-k2.6/astrazeneca   &  97 & $65{:}46 \pm 39{:}54$ & 98\% \\
\bottomrule
\end{tabular}

\end{table}


\section{Discussion}

This work reframes NMR elucidation for small organic molecules as an agentic, constraint-driven search problem rather than an end-to-end model that maps spectra to structures. Our results support this hypothesis: an agent with access to a curated environment can achieve competitive performance against models trained on large simulated datasets. It achieves a 80\% top-1 accuracy on the Chemistry Education dataset, performs on par with graduate-level chemistry students on Alberts, and reaches 19\% top-1 accuracy on the more challenging AstraZeneca dataset, which contains complex drug-like molecules measured under laboratory conditions. The failure cases show where the current approach remains limited. Some errors occur when the agent commits too early to a molecular family and cannot revise it later, especially when spectra contain noise, impurities, artifacts, or overlapping peaks. As molecular complexity increases, the reasoning traces become longer and harder to navigate. Robustness across repeated independent runs is another important limitation: some tasks converge consistently to the same answer, while others do not. Understanding this variability and quantifying uncertainty are important future directions.

A useful aspect of the agentic approach is that it can scale by improving the environment around the model: more reliable preprocessing, stronger verification checks, better search strategies, and additional modalities such as IR, mass spectrometry, EPR, UV-vis, microwave, Raman, and 2D NMR. Overall, these results suggest that agentic puzzle solving is a promising framework for small-molecule NMR structure elucidation. Going forward, we are extending the framework into a multi-agent system that can integrate additional modalities while improving context management. This framework should be viewed as a supporting tool for chemists, and not as a substitution; it can generate candidates, expose its reasoning, and help prioritize structures for further experimental validation.

\section*{Acknowledgments}
We would like to acknowledge the support of the Simons Foundation and of
Schmidt Sciences. This work was supported in part by the AI2050 program at Schmidt Sciences
(Grant G-25-70028). Additionally, computations were run at facilities supported by the Scientific Computing Core at the Flatiron Institute. The Flatiron Institute is a division of the Simons Foundation.

\pagebreak
\bibliography{references}
\bibliographystyle{icml2026}

\clearpage

\appendix
\onecolumn

\section*{Supplementary Material}
\label{supp:examples}
See extensive examples of correct, incorrect, near misses and mixed predictions by our agent. Combination of all three datasets, with Alberts and AstraZeneca being ran for three independent runs.

\vspace*{2cm}

\begin{center}

\noindent\makebox[\textwidth][c]{%
\includegraphics[width=0.88\textwidth, trim=0 170 0 0, clip]{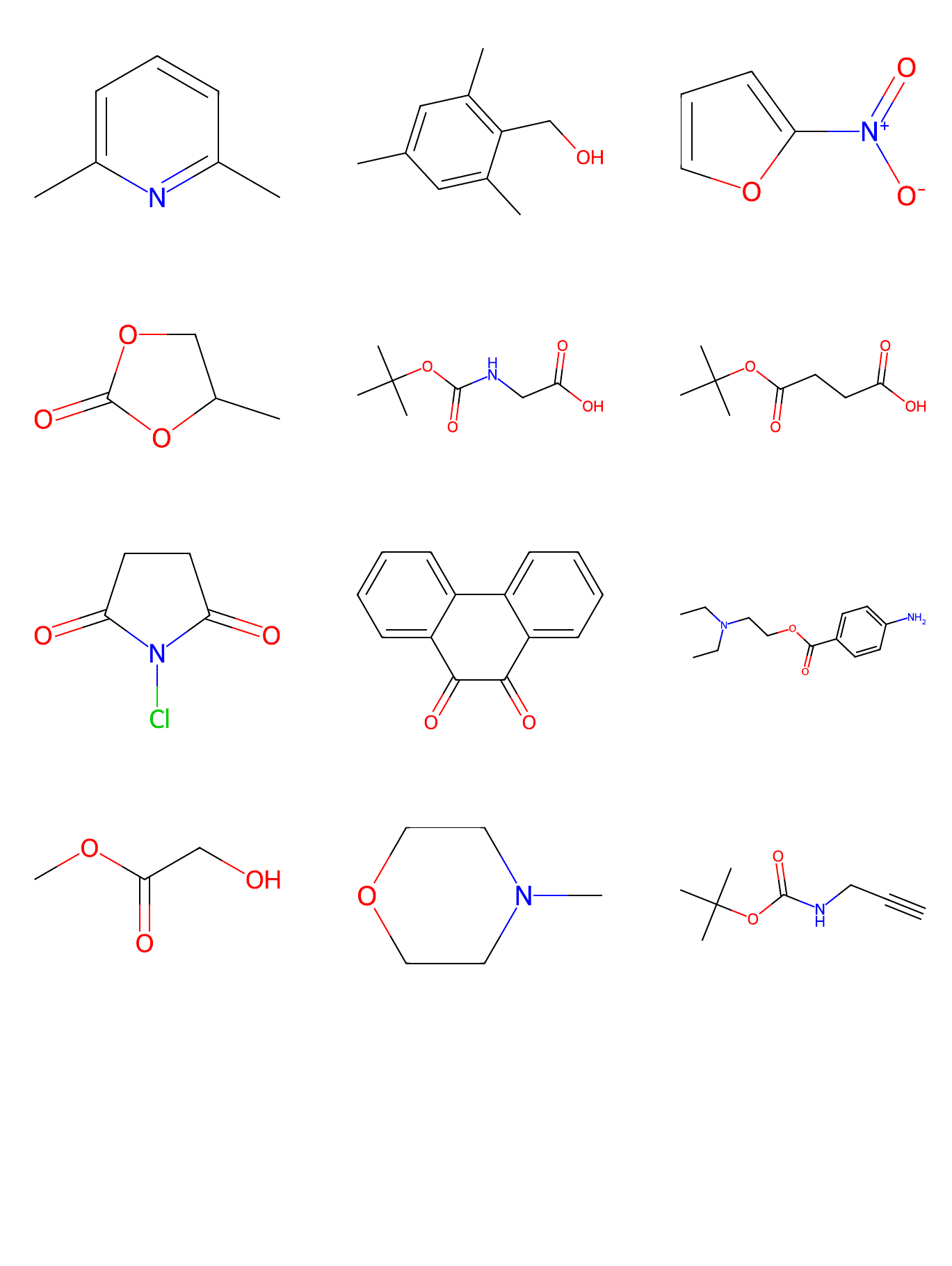}%
}

\captionof{figure}{\textbf{Additional correct predictions by our agent.} Examples of correct top-1 predictions from \texttt{kimi-k2.6}.}

\label{fig:supp_correct_examples}

\end{center}

\clearpage

\vspace*{2cm}

\begin{center}

\noindent\makebox[\textwidth][c]{%
\includegraphics[width=0.88\textwidth, trim=0 170 0 0, clip]{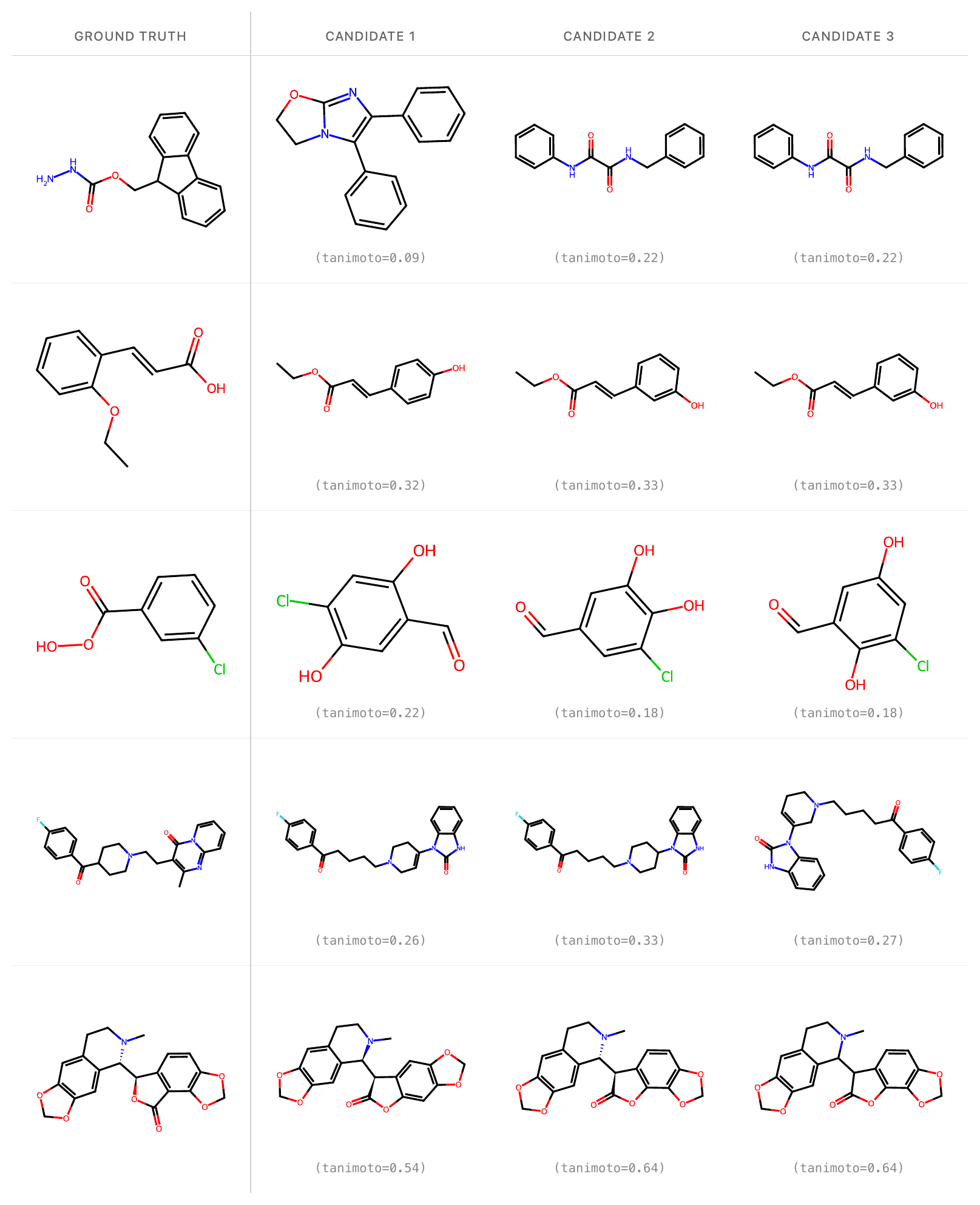}%
}

\captionof{figure}{\textbf{Additional incorrect predictions by our agent.} Representative failure cases from \texttt{kimi-k2.6}.}

\label{fig:supp_simple_examples}

\end{center}

\clearpage

\vspace*{2cm}

\begin{center}

\noindent\makebox[\textwidth][c]{%
\includegraphics[width=0.88\textwidth, trim=0 170 0 0, clip]{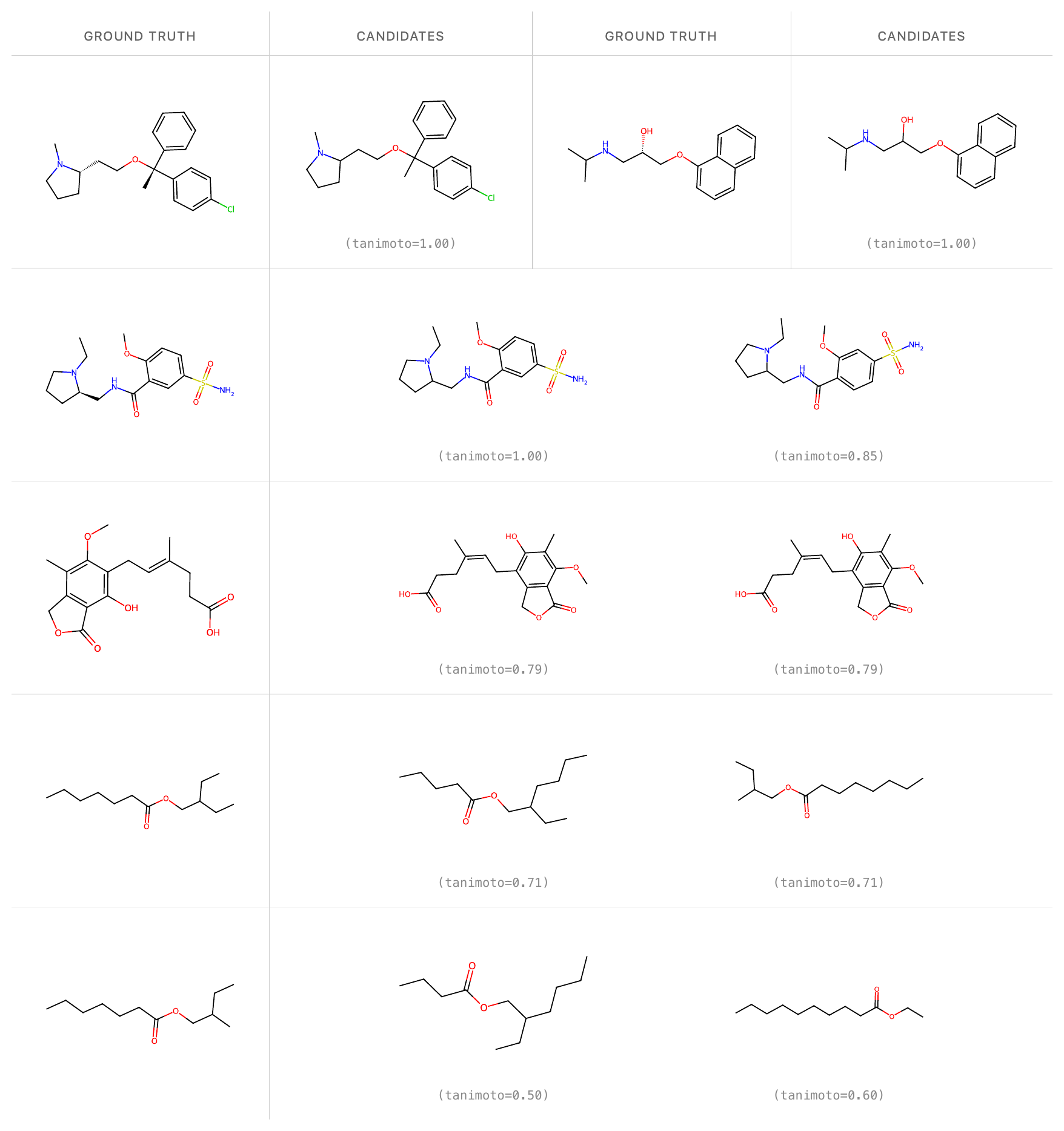}%
}

\captionof{figure}{\textbf{Additional near-miss predictions by our agent.} Representative near-miss cases from \texttt{kimi-k2.6}.}

\label{fig:supp_near_miss_examples}

\end{center}

\clearpage

\vspace*{2cm}

\begin{center}

\noindent\makebox[\textwidth][c]{%
\includegraphics[width=0.88\textwidth, trim=0 170 0 0, clip]{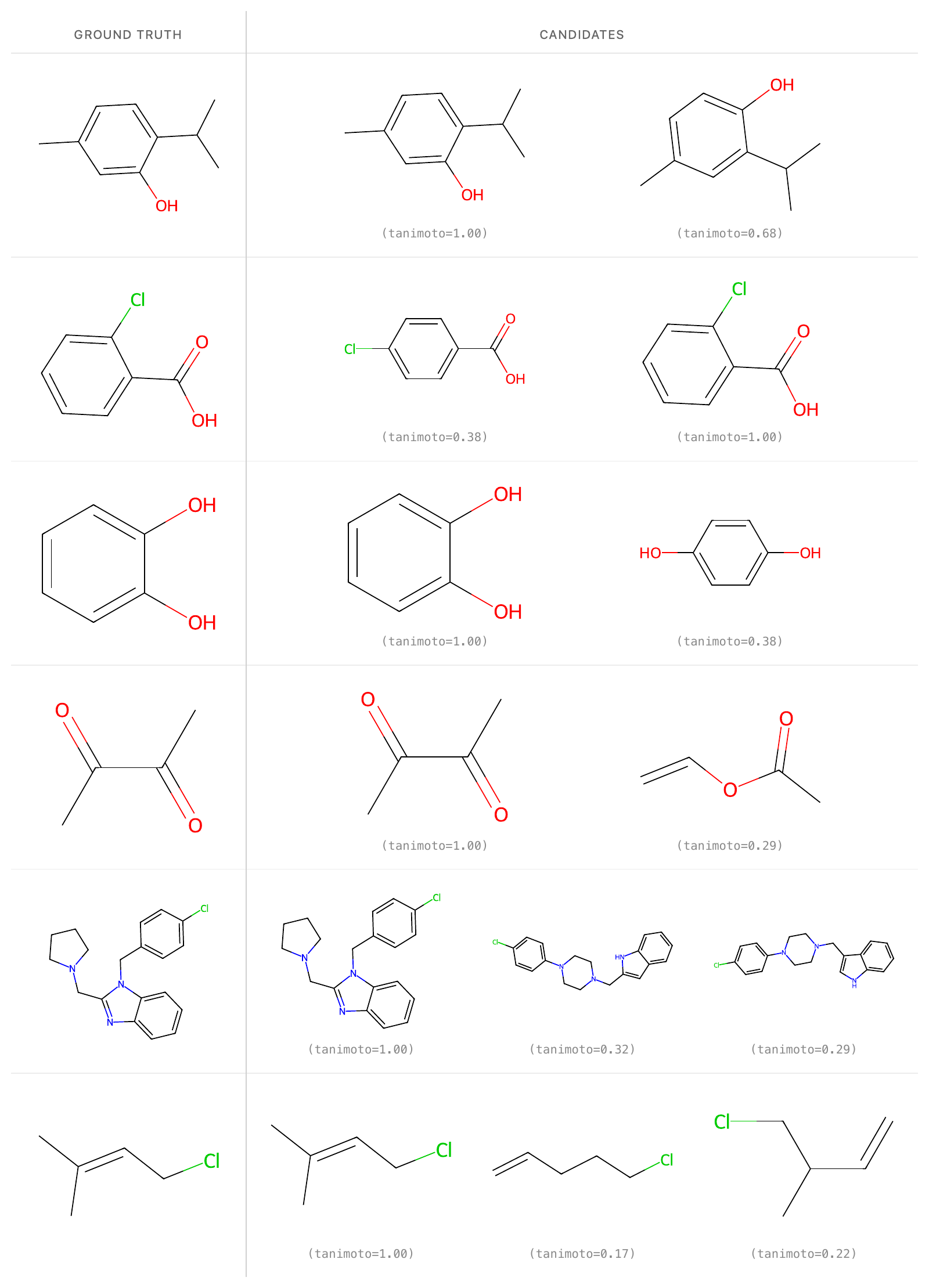}%
}

\captionof{figure}{\textbf{Additional mixed predictions by our agent.} Representative cases where the agent predicts the correct answer in some, but not all, independent runs from \texttt{kimi-k2.6}.}

\label{fig:supp_mixed_examples}

\end{center}

\clearpage

\section*{Reasoning Traces}
\label{supp:reasoning_traces}
The excerpts below are reproduced verbatim from the agent's reasoning traces.
They are grouped into the reasoning skills the agent applies during elucidation
and the reasoning failures we observed, and are meant to be illustrative rather
than exhaustive.

Each excerpt illustrates a distinct operator from the manual
structure-elucidation workflow.

\textbf{Reasoning skill: Candidate enumeration and elimination.} In a trace from ChemEdu
(2,4,6-tribromoaniline), the agent iterates over candidate structures consistent
with the molecular formula \ce{C6H4Br3N}, pruning them against both atom counts
and the observed $^{13}$C chemical shifts, and reasons about whether anomalous
peaks could be spectral artifacts.

\begin{tcolorbox}[
  breakable,
  colback=gray!5,
  colframe=gray!50!black,
  title={ChemEdu Task: 2,4,6-Tribromoaniline (\ce{C6H4Br3N}), Kimi 2.6},
  fonttitle=\bfseries,
  left=2mm, right=2mm, top=1mm, bottom=1mm,
]
What about a pyridine with a bromomethyl group?
2,6-dibromo-3-(bromomethyl)pyridine: \ce{C6H4Br3N}. Yes!

This compound would have:
\begin{itemize}\itemsep0pt
  \item A pyridine ring with 2 bromines and 1 bromomethyl group
  \item 2 aromatic protons
  \item 2 aliphatic protons (\ce{CH2Br})
\end{itemize}

The $^{1}$H NMR would show:
\begin{itemize}\itemsep0pt
  \item 2 aromatic protons
  \item 1 \ce{CH2} signal around 4.5 ppm
\end{itemize}

The $^{13}$C NMR would show:
\begin{itemize}\itemsep0pt
  \item 5 aromatic carbons (pyridine ring)
  \item 1 aliphatic carbon (\ce{CH2Br}) around 30--40 ppm
\end{itemize}

This doesn't match the observed $^{13}$C peaks at 82 ppm.

What about 2,4-dibromo-3-(bromomethyl)pyridine? Same issue.

What about a compound with a \ce{CBr3} group? Like \ce{CBr3-CH=CH-NH2} or
something?

\ce{CBr3} is the tribromomethyl group. If we have \ce{CBr3-C2H2-N}, that would
be \ce{C3H2Br3N}. We need \ce{C6H4Br3N}.

So we need 3 more carbons and 2 more hydrogens. Maybe \ce{CBr3-C6H4-NH2}? That
would be \ce{C7H6Br3N}. Too many carbons.

Or \ce{CBr3-C5H4N}? That's \ce{C6H4Br3N}. This would be
2-(tribromomethyl)pyridine or 3-(tribromomethyl)pyridine or
4-(tribromomethyl)pyridine.

In 2-(tribromomethyl)pyridine:
\begin{itemize}\itemsep0pt
  \item The \ce{CBr3} carbon would appear around 50--60 ppm (similar to
        \ce{CBr4} at ${\sim}{-30}$ ppm, but attached to pyridine it might be
        different)
  \item The pyridine ring carbons would be around 120--150 ppm
\end{itemize}

The $^{13}$C at 82 ppm doesn't match the \ce{CBr3} carbon.

What about a compound with a \ce{CBr2} group? Like \ce{CHBr2-C5H4N}? That's
\ce{C6H5Br2N}. Not \ce{C6H4Br3N}.

I'm running out of ideas for non-aniline structures with formula \ce{C6H4Br3N}.

Let me go back to the tribromoaniline hypothesis and try to make sense of the
82 ppm peaks.

Actually, I just realized something. In some NMR spectra, especially if the
sample is not properly prepared, the baseline can have artifacts that look like
peaks. But three distinct peaks at 82 ppm in a regular pattern seems unlikely to
be baseline artifacts.

Another thought: could the 82 ppm peaks be from a spinning sideband? In
solid-state NMR, spinning sidebands appear at regular intervals from the main
peak. But this is a liquid-state NMR (solution NMR), so spinning sidebands
shouldn't be an issue.
\end{tcolorbox}

\textbf{Reasoning Skill: Spectral assignment and solvent reconciliation.} In a trace from
AstraZeneca (diltiazem), the agent groups the observed $^{13}$C peaks by
chemical-shift region, proposes a tentative assignment for each group, and
reconciles the 29 observed peaks against the 22 carbons in the molecular formula
by attributing signals near 40~ppm to the DMSO-$d_6$ solvent.

\begin{tcolorbox}[
  breakable,
  colback=gray!5,
  colframe=gray!50!black,
  title={AstraZeneca Task: Diltiazem, (2S,3S)-cis isomer (\ce{C22H26N2O4S}), Kimi 2.6},
  fonttitle=\bfseries,
  left=2mm, right=2mm, top=1mm, bottom=1mm,
]
The $^{13}$C NMR shows:
\begin{itemize}\itemsep0pt
  \item 169.69, 166.66 --- two carbonyls (ester/amide)
  \item 145.50, 135.38, 132.31, 131.63, 131.32, 129.82, 127.90, 127.38, 126.82, 125.19, 124.36, 113.84 --- 12 aromatic/olefinic carbons
  \item 70.76 --- \ce{C-O}
  \item 56.82, 56.23, 55.48, 53.64 --- four carbons, likely \ce{C-N} or \ce{C-O}
  \item 48.32, 47.17, 45.56 --- three carbons, likely \ce{C-N}
  \item 40.60, 40.12, 39.70, 39.28 --- four carbons, likely near S or N
  \item 22.79, 20.64, 19.80 --- three methyl/methylene carbons
\end{itemize}

Total 29 $^{13}$C peaks. But the formula has only 22 carbons. This means some
peaks are from solvent (DMSO at ${\sim}40$~ppm) and possibly the compound has
symmetry.

Wait, DMSO-$d_6$ has a $^{13}$C peak at ${\sim}39.5$~ppm. Looking at the peaks
around 40~ppm:
\begin{itemize}\itemsep0pt
  \item 40.60, 40.12, 39.70, 39.28
\end{itemize}

These could include the DMSO peak at 39.42 (as noted in the solvent peak). So
40.60, 40.12 might be from the compound, and 39.70, 39.28 might be
DMSO-related.
\end{tcolorbox}

\textbf{Reasoning skill: Symmetry-based isomer discrimination.} In a trace from Alberts
(phenanthrene-9,10-dione), the agent reasons about molecular symmetry to
discriminate between two isomeric quinones (anthraquinone vs.\
phenanthrene-9,10-dione) that share the formula \ce{C14H8O2}, by counting the
unique $^{13}$C environments implied by each candidate's symmetry and
cross-checking against the $^{1}$H pattern.

\begin{tcolorbox}[
  breakable,
  colback=gray!5,
  colframe=gray!50!black,
  title={Alberts Task: 9,10-Phenanthrenequinone (\ce{C14H8O2}), GPT 5.4},
  fonttitle=\bfseries,
  left=2mm, right=2mm, top=1mm, bottom=1mm,
]
\textbf{Evaluating molecular symmetry}

I'm considering the symmetry of $C_2$, which may involve 4 sets of 2H. I need to
compare the $^{13}$C count, possibly with 7 unique carbons. For
phenanthrene-9,10-dione, it has a carbonyl in the central ring. With $C_2$
symmetry, there might actually be 7 unique carbons: one carbonyl, two bridgehead
quaternary, and four aromatic CHs. Anthraquinone and phenanthrenequinone share
the formula \ce{C14H8O2}, so I should also compare their $^{1}$H patterns for
discrimination.

\textbf{Examining anthraquinone structure}

I'm looking at anthraquinone and considering the possibility of having 5 unique
aromatic signals. It seems plausible if I have two quaternary and three CH
groups. However, I notice that $^{1}$H shows four unique protons, indicating
four distinct CH carbons, which might not match up if there's some signal
overlap. I'm trying to map out its symmetry, particularly how carbonyls and
protons are arranged. There might only be two distinct proton environments based
on symmetry analysis.

\textbf{Evaluating symmetries in compounds}

I'm examining a mirror plane along the long axis through specific carbons,
swapping top and bottom. I see two proton sets of 4H each! When looking at known
anthraquinone data, it aligns with proton shifts. However, the multiplet centers
show a range that makes me reconsider; it might not be anthraquinone but rather
phenanthrenequinone due to symmetry differences. I'll have to verify the unique
$^{13}$C count for phenanthrene-9,10-dione to confirm.
\end{tcolorbox}

\textbf{Reasoning skill: Tool-grounded verification.} In another trace from Alberts
(Boc-propargylamine), the agent proposes candidate SMILES strings for the formula
\ce{C8H13NO2} and issues an explicit call to the \texttt{Verify\_smiles} tool to
confirm the atom counts, using a spectral feature (the methyl doublet) to flag a
formula-valid candidate as inconsistent. Tool calls are typeset in a monospace
font to distinguish them from the agent's reasoning.

\begin{tcolorbox}[
  breakable,
  colback=gray!5,
  colframe=gray!50!black,
  title={Alberts Task: tert-Butyl N-(prop-2-yn-1-yl)carbamate (\ce{C8H13NO2}), GPT 5.4},
  fonttitle=\bfseries,
  left=2mm, right=2mm, top=1mm, bottom=1mm,
]
\textbf{Checking candidate formulas}

I'm considering a structure that might include a \ce{CH3} group, which leads me
to the formula \texttt{N\#CC(C)C(=O)OC(C)(C)C}, potentially giving me \ce{C8}. I
think it's important to confirm that it matches \ce{C8H13NO2}. After counting the
atoms, it seems I do have the right totals: one nitrile, one alpha carbon, one
methyl, one carbonyl, and four from tert-butyl. I'm considering calling this
candidate tert-butyl 2-cyano-2-methylacetate, but it does bring up some
inconsistencies with the methyl doublet.

\toolcall{Verify\_smiles}

\textbf{Verifying chemical formulas}

I'm looking at the chemical structure of Boc-propargylamine with the formula
\texttt{CC(C)(C)OC(=O)NCC\#C}. It seems the count formula checks out. Now, I want
to compare it to O-propargyl tert-butylcarbamate, which has a formula of
\texttt{C\#CCOC(=O)NC(C)(C)C}. I'm thinking the count might be similar, so I'll
need to verify that next. Let's ensure everything aligns correctly before moving
forward.
\end{tcolorbox}

\textbf{Reasoning failure: Anchoring on a disconfirmed core.} A failure case from the AstraZeneca
dataset, where the agent anchors on an incorrect structural family and explains
away each piece of disconfirming evidence rather than abandoning it. The excerpts
below are reproduced verbatim from non-contiguous points in the reasoning trace;
omitted spans are marked \texttt{[\ldots]}.

\begin{tcolorbox}[
  breakable,
  colback=gray!5,
  colframe=gray!50!black,
  title={AstraZeneca Task: 9H-Fluoren-9-ylmethyl carbazate (\ce{C15H14N2O2}), GPT 5.4},
  fonttitle=\bfseries,
  left=2mm, right=2mm, top=1mm, bottom=1mm,
]
I'm wondering if 143.78 and 141.47 could be oxazoline C=N carbons. But I realized
that C=N in oxazolines typically appears around 161-165, so 143 seems too low. It
could be that these are aryl C-O/N adjacent to nitrogens, rather than imines.
Then I consider 2,2'-methylene bis(oxazoline) derivatives\ldots

\smallskip\centerline{\ttfamily[\ldots]}\smallskip

I thought of N-methylbenzimidazole, but it falls short on carbons by 4.

\smallskip\centerline{\ttfamily[\ldots]}\smallskip

However, it seems N-methylbenzimidazole's NCH3 isn't aligning with my
expectations. I need to keep refining this!

\smallskip\centerline{\ttfamily[\ldots]}\smallskip

adding 1,3-dioxolane and N-methyl results in C11H14N2O2, not 15, which raises
questions. So, I might need 4 more carbons, perhaps by considering a phenyl
substituent\ldots

\smallskip\centerline{\ttfamily[\ldots]}\smallskip

I'm thinking about the formula file. It seems unlikely to be wrong, but I wonder
if maybe 3.75 could indicate bridge CH2 and 6.02 an impurity. I should explore
potential formulas for o-phenylenebisoxazoline derivatives, looking into options
like a methylene-substituted ring or benzimidazole.
\end{tcolorbox}

\textbf{Reasoning failure: A missed shift diagnostic in a near-miss.} A
near-miss from the AstraZeneca dataset on fenbufen (\ce{C16H14O3}). The agent
recovers all the correct fragments---a monosubstituted phenyl, a
para-disubstituted ring, a \ce{-CH2CH2-} linker, a ketone at 198~ppm, and an
acid/ester carbon at 174~ppm---but assembles them with the carboxylic acid
placed directly on an aromatic ring (a benzoic acid) rather than at the end of
the aliphatic chain. The correct connectivity is in fact present in its own
candidate list (submitted as candidate~2). The single clue that distinguishes
the two is the shift of the acid carbon: 174~ppm is diagnostic of an
\emph{aliphatic} carboxylic acid (\ce{R-CH2-COOH}, ${\sim}174$--180~ppm), not an
aromatic one (\ce{Ar-COOH}, ${\sim}167$--172~ppm). The agent observed this peak
but classified it only with a coarse 170--175~ppm range that lumps the two acid
types together, so it never used the diagnostic to eliminate its own
benzoic-acid answer; it then discarded the correct isomer for an unrelated and
weaker reason. Excerpts are reproduced verbatim from non-contiguous points in
the trace; omitted spans are marked \texttt{[\ldots]}.

\begin{tcolorbox}[
  breakable,
  colback=gray!5,
  colframe=gray!50!black,
  title={AstraZeneca Task: Fenbufen (\ce{C16H14O3}), Kimi 2.6},
  fonttitle=\bfseries,
  left=2mm, right=2mm, top=1mm, bottom=1mm,
]
Let me reconsider the 174 ppm carbonyl. The evidence table says it has no
candidate groups. 174 ppm is typical for:
\begin{itemize}\itemsep0pt
  \item Ester: 170--175 ppm
  \item Carboxylic acid: 170--175 ppm
  \item Amide: 165--175 ppm
  \item Lactone: 165--175 ppm
  \item $\alpha,\beta$-unsaturated ketone: lower, ${\sim}190$--200
\end{itemize}

So 174 is likely an ester or carboxylic acid.

\smallskip\centerline{\ttfamily[\ldots]}\smallskip

For isomer 2: The aromatic pattern would be different --- a biphenyl system
would have more complex aromatic signals.

\smallskip\centerline{\ttfamily[\ldots]}\smallskip

This pattern (2H, 2H, 2H, 2H, 1H) = 9H aromatic is consistent with:
\begin{itemize}\itemsep0pt
  \item A para-disubstituted benzene (4H, appearing as 2 doublets)
  \item A monosubstituted benzene (5H, appearing as 2H + 2H + 1H)
\end{itemize}

This strongly supports isomer 1: \texttt{O=C(O)c1ccc(CCC(=O)c2ccccc2)cc1}

\smallskip\centerline{\ttfamily[\ldots]}\smallskip

The structure \texttt{O=C(O)c1ccc(CCC(=O)c2ccccc2)cc1} fits all these features
perfectly. Let me submit this as the final answer.
\end{tcolorbox}

\textbf{Reasoning failure: Locally-reasonable chemistry on globally-broken
data.} A failure from the Alberts dataset on \emph{meta}-chloroperoxybenzoic
acid (mCPBA, \ce{C7H5ClO3}), notable because both backbones we tested converge
on the \emph{same} wrong structure, a chlorodihydroxybenzaldehyde. The trap is
that this incorrect structure shares the exact molecular formula and degree of
unsaturation ($\mathrm{DoU}=5$) of mCPBA, so neither the formula nor the DoU
check can separate them. The one feature that would betray the peroxyacid---a
carbonyl carbon near 167~ppm---is absent from the preprocessed $^{13}$C data:
after solvent removal only two carbons survive (a spurious peak at 151~ppm and
one at 131~ppm), and the true carbonyl is missing entirely. With no positive
evidence for the exotic \ce{-C(=O)-O-O-H} group, and an evidence table that
offers aldehyde- and phenol-like assignments for the surviving peaks,
``aldehyde plus two phenols'' becomes the most shift-plausible reading of the
degraded spectrum. The agent then performs locally-reasonable chemistry on
globally-corrupted data; because both models receive the identical broken
spectrum and the same evidence-table hints, neither can recover and both fail in
the same way. Excerpts (\texttt{gpt-5.4}) are reproduced verbatim from
non-contiguous points in the trace; omitted spans are marked \texttt{[\ldots]}.

\begin{tcolorbox}[
  breakable,
  colback=gray!5,
  colframe=gray!50!black,
  title={Alberts Task: \textit{meta}-Chloroperoxybenzoic acid (mCPBA, \ce{C7H5ClO3}), GPT 5.4},
  fonttitle=\bfseries,
  left=2mm, right=2mm, top=1mm, bottom=1mm,
]
The values suggest it could be chlorosalicylaldehyde or chlorohydroxybenzoic
acid. The presence of 1.68 ppm indicates a possible water impurity, while the
remaining non-solvent peaks could indicate specific functional groups.

\smallskip\centerline{\ttfamily[\ldots]}\smallskip

(evidence table, $^{13}$C rows after solvent exclusion)
\begin{itemize}\itemsep0pt
  \item $^{13}$C 151.136 --- Oxygenated aromatic carbon(s), likely Ar--C--OH
  \item $^{13}$C 131.190 --- Aromatic sp$^2$ carbon attached to H / substituted
        aryl C
\end{itemize}

\smallskip\centerline{\ttfamily[\ldots]}\smallskip

Interpretation: the formula requires one Cl and three O atoms. The 9.74 ppm
singlet strongly supports an aldehyde. The remaining O atoms are most
economically explained as two phenols. Thus the core is a
chlorodihydroxybenzaldehyde.

\smallskip\centerline{\ttfamily[\ldots]}\smallskip

Chosen structure: \texttt{O=Cc1cc(O)c(Cl)c(O)c1} =
4-chloro-3,5-dihydroxybenzaldehyde. \ldots The proton-count failure is not
disqualifying here because the raw integral sum is inflated by residual
\ce{CHCl3} and water peaks that are plainly present in the processed data.
\end{tcolorbox}

\end{document}